\documentclass[letterpaper, 10 pt, conference]{ieeeconf}  

\IEEEoverridecommandlockouts                              

\usepackage{graphicx} 
\usepackage{amsmath} 
\usepackage{amssymb}  

\usepackage{hyperref}
\usepackage{booktabs}
\usepackage{diagbox}
\usepackage{multirow}

\title{\LARGE \bf
Learning to Drive (L2D) as a Low-Cost Benchmark for\\ Real-World Reinforcement Learning
}

\author{Ari Viitala$^{*}$, Rinu Boney$^{*}$, Yi Zhao, Alexander Ilin, and Juho Kannala
\thanks{$^{*}$ Equal contribution}%
\thanks{The authors are with Department of Computer Science, Aalto University, Finland.
        {\tt\small firstname.lastname@aalto.fi}}%
}

\begin{document}

\maketitle
\thispagestyle{empty}
\pagestyle{empty}

\begin{abstract}
We present Learning to Drive (L2D), a low-cost benchmark for real-world reinforcement learning (RL). L2D involves a simple and reproducible experimental setup where an RL agent has to learn to drive a Donkey car around three miniature tracks, given only monocular image observations and speed of the car. The agent has to learn to drive from disengagements, which occurs when it drives off the track.
We present and open-source our training pipeline, which makes it straightforward to apply any existing RL algorithm to the task of autonomous driving with a Donkey car. 
We test imitation learning, state-of-the-art model-free, and model-based algorithms on the proposed L2D benchmark. Our results show that existing RL algorithms can learn to drive the car from scratch in less than five minutes of interaction.
We demonstrate that RL algorithms can learn from sparse and noisy disengagement to drive even faster than imitation learning and a human operator.
\end{abstract}

\section{Introduction}

Deep reinforcement learning (RL) algorithms are often criticized to be sample-inefficient, highly sensitive to hyperparameters, and hard to reproduce \cite{duan2016benchmarking, islam2017reproducibility, henderson2018deep, mania2018simple}. Recently, there has been significant progress towards sample-efficient and robust deep RL algorithms \cite{haarnoja2018softb, hafner2020dream}, largely due to simulator benchmarks \cite{brockman2016openai, tassa2018deepmind}. It is important to ground this progress by measuring their performance in real-world tasks.

The dominating approach in real-world reinforcement learning is heavily based on the use of simulators: RL agents are first trained in simulated environments and then adapted to the real world \cite{akkaya2019solving, bewley2019learning}. This approach has limitations as 1)~one has to build a (complex) simulator of the real world and 2)~a policy trained on the simulator can easily fail in the real world as simulators lack many complexities of reality. This approach is somewhat similar to feature engineering in tackling a machine learning problem: designing a good simulator, just like designing a good set of features for complex machine learning problems, may take years of research.

The alternative approach is to train an RL agent by interacting directly with the real world. The difficulty (and the potential advantage) of this approach is that the agent must cope with all the complexities of reality during training. This approach can be seen analogous to learning a good set of features via representation learning \cite{goodfellow2016} as opposed to feature engineering to solve a machine learning problem. The progress in this direction can be boosted by standardized and broadly accessible physical platforms for RL, just like standard datasets \cite{imagenet} accelerated research in machine learning.

Recently, there have been several prominent demonstrations of deep RL in real-world tasks. One exciting application is autonomous driving, in which the first application of deep RL was demonstrated by \cite{kendall2019learning}. However, reproducing those results is difficult for a broad research community as it was demonstrated using a full-sized car.

\begin{figure}[t]
\centering
\includegraphics[width=\linewidth]{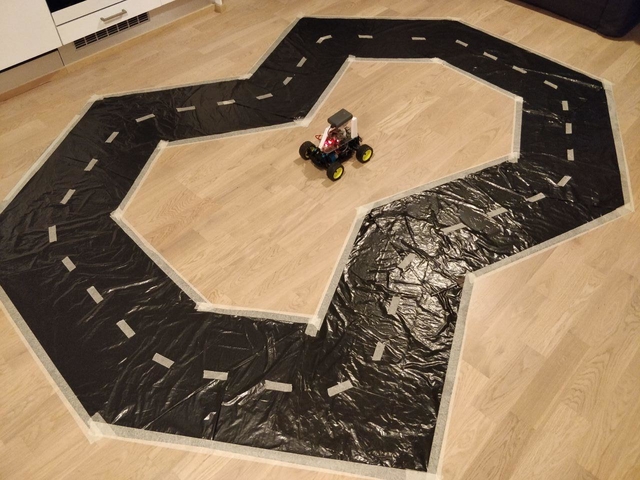}
\caption{Our Donkey car and one of the (${\sim}7$m long) tracks used in our real-world experiments. The car starts from the same position in all episodes. The episode terminates if the car drives off the track. An episode is successful if the car completes three laps around the track.}
\label{fig:track-photo}
\end{figure}

In this paper, we propose L2D (Learning to Drive), a low-cost real-world benchmark in which the RL agent learns how to drive using human disengagements as rewards. We adapt an open-source self-driving platform called Donkey car \cite{donkeycar} which requires minimal hardware costing $\$250$ and takes two hours to assemble. We comprehensively describe the experimental setup to make the proposed real-world platform reproducible exactly.  We provide all the details of the experimental settings including the track dimensions and our training trajectories. We open-source the required software pipeline to make it possible to use any RL algorithm on the Donkey car platform.

The proposed L2D benchmark is fun to play with and has many challenges of real-world systems \cite{dulac2019challenges}. 1)~High-dimensional observations: The agent has to learn to control the car directly from a sequence of images captured by an on-board camera. 2)~Delayed observations, actions, and rewards: Unlike in simulators, the agent has to deal with inevitable (stochastic) delays in the real system. 3)~Non-stationary dynamics: The throttle control of the car is highly sensitive and is adversely affected by factors such as the friction of the surface, battery voltage, and temperature. 4)~Real-time inference: The policy has to be able to run in real-time at the control frequency of the system. 5)~Data-efficiency: The agent must learn from limited samples on the real car.

We provide results of three popular RL algorithms for the proposed benchmark: 1)~an imitation learning agent, 2)~a model-free RL algorithm \cite{yarats2019improving} combining variational autoencoder (VAE) and soft-actor-critic (SAC) and 3)~a model-based RL algorithm called Dreamer \cite{hafner2020dream}. This paper, to the best of our knowledge, provides the first successful demonstration of Dreamer in a real-world task.

\begin{figure}[t]
\centering
\vspace{3mm}
\includegraphics[height=0.6\linewidth,trim={0 0 0 10mm},clip]{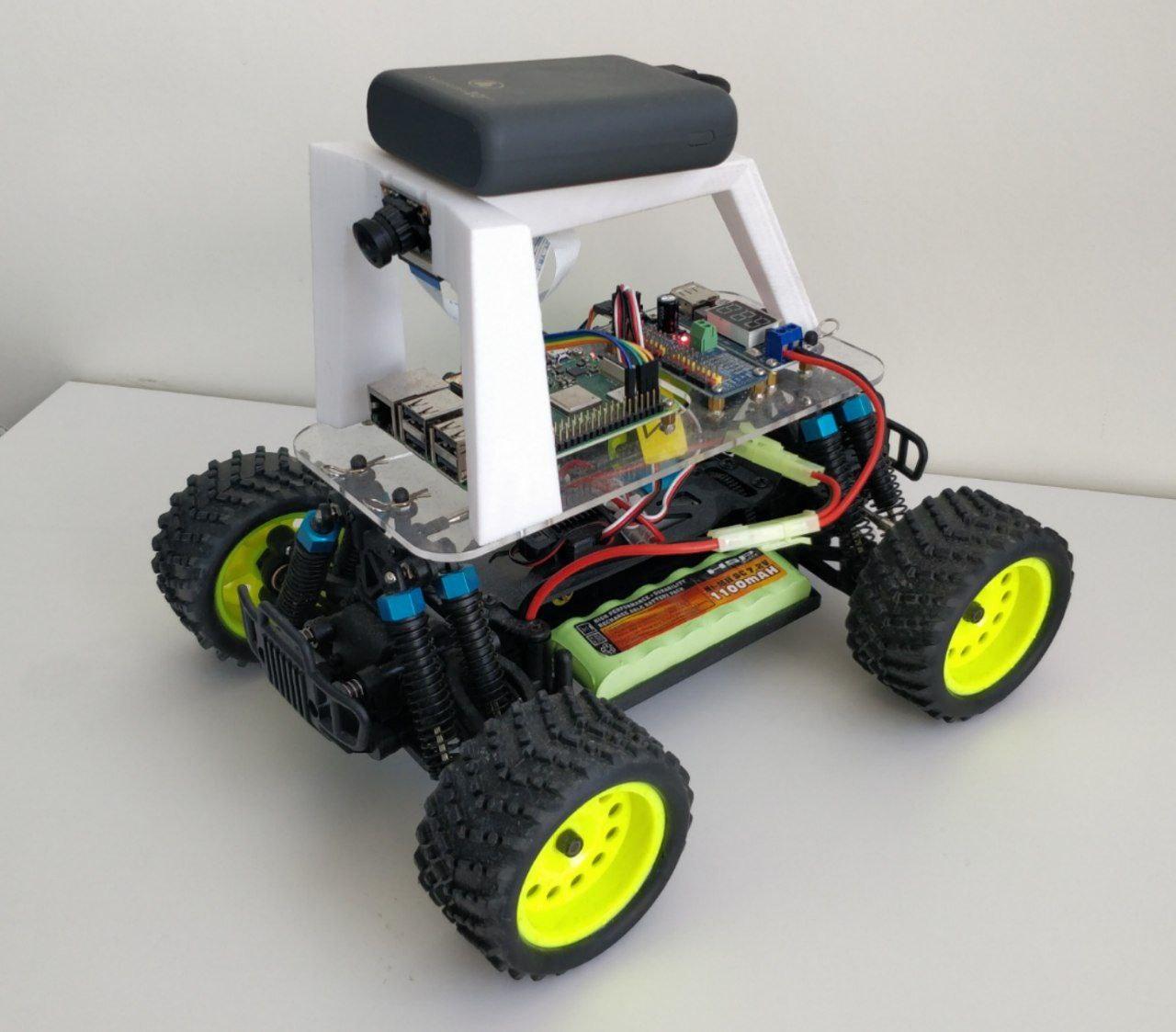}
\caption{Our Donkey car. The back wheels are throttled using a brushless motor and the front wheels are steered using a servo. The Raspberry Pi 4, a camera, and other supporting hardware are mounted on a 3D printed cage.}
\label{fig:donkey-car}
\end{figure}

\section{Related Work}


The need for standardized physical benchmarks for RL has been gaining attraction recently. ROBEL \cite{ahn2020robel} is a recently introduced open-source platform for benchmarking in robot learning, consisting of a four-legged robot (costing \$4200) and a three-fingered hand robot (costing \$3500). While ROBEL is aimed at research on learning challenging locomotion and manipulation tasks from low-dimensional state spaces, we focus on RL from high-dimensional observations. REPLAB \cite{yang2019replab} is a reproducible and self-contained platform (costing \$2000) with a low-cost arm for vision-based manipulation tasks. Wheeled robot platforms such as the Donkey car used in this paper are significantly cheaper (in the range \$200--\$500) and requires much less maintenance. 
Amazon DeepRacer \cite{balaji2019deepracer} is an autonomous racing platform for experimentation in sim2real RL. We use the completely open-source Donkey car platform and focus on the direct application of RL in the real world. Building high-fidelity simulators that allow for sim2real is non-trivial, for example, we found driving in the Donkey car simulator to be much simpler than the physical platform (due to simpler dynamics and absence of non-stationarities and delays).
Other available wheeled platforms include DuckieBot \cite{paull2017duckietown}, MuSHR \cite{gorjup2020mushr}, DJI Robotmaster S1 \cite{robomaster}, and Nvidia JetBot \cite{jetbot}.
While there exist multiple wheeled platforms, we adapt one for repeatable and reproducible research in real-world RL for the first time.

CARLA \cite{dosovitskiy2017carla} is an open-source simulator for benchmarking in autonomous driving research. CARLA aims to evaluate autonomous driving algorithms in controlled scenarios of increasing difficulty. In this work, we do not attempt to solve autonomous driving but focus on grounding the progress of RL research in a real-world task related to autonomous driving.


There exist many papers demonstrating the use of reinforcement learning on real-world tasks.
Perhaps the most related work is \cite{kendall2019learning} in which they demonstrated sample-efficient RL to drive a full-sized car from just monocular image observations.
Sample-efficient RL has also been successfully demonstrated on
various locomotion \cite{singh2019, ha2020learning,yang2020data} and manipulation tasks \cite{haarnoja2018learning,singh2019,zhu2019ingredients}.
Most of these demonstrations were done on expensive robots. In contrast, our benchmark is low-cost, easily accessible, and reproducible.


Many real-world RL agents were based on model-free learning: the DDPG algorithm \cite{lillicrap2016continuous} was used in \cite{kendall2019learning,zhu2019ingredients}, while \cite{haarnoja2018learning, singh2019, ha2020learning} adopted soft actor-critic (SAC) \cite{haarnoja2018softa, haarnoja2018softb}.
High-dimensional observations were often compressed by means of a pre-trained autoencoder \cite{shelhamer2016loss, higgins2017darla, nair2018visual}, which was first proposed to assist RL in \cite{lange2010deep}.
More recent works \cite{tassa2018deepmind, yarats2019improving} demonstrated stable joint representation learning by truncating the gradients from the actor loss to the encoder. Our experiments demonstrate stable joint training of the autoencoder and the RL agent on the L2D benchmark.

A few recent papers demonstrated the effectiveness of model-based RL for learning how to fly a drone \cite{lambert2019low,becker2020learning} and for legged locomotion \cite{yang2020data}. There, the world models were built from relatively low-dimensional observation signals. The effective use of world models trained from high-dimensional observations for real-world reinforcement learning tasks were demonstrated in 
\cite{ebert2018visual,zhang2019solar}.
In this paper, we use the state-of-the-art model-based RL algorithm called Dreamer \cite{hafner2020dream} and demonstrate it for the first time on a real-world task.

\section{Benchmark}
\label{sec:setup}

\begin{figure*}[t]
\centering
\includegraphics[width=1.0\linewidth]{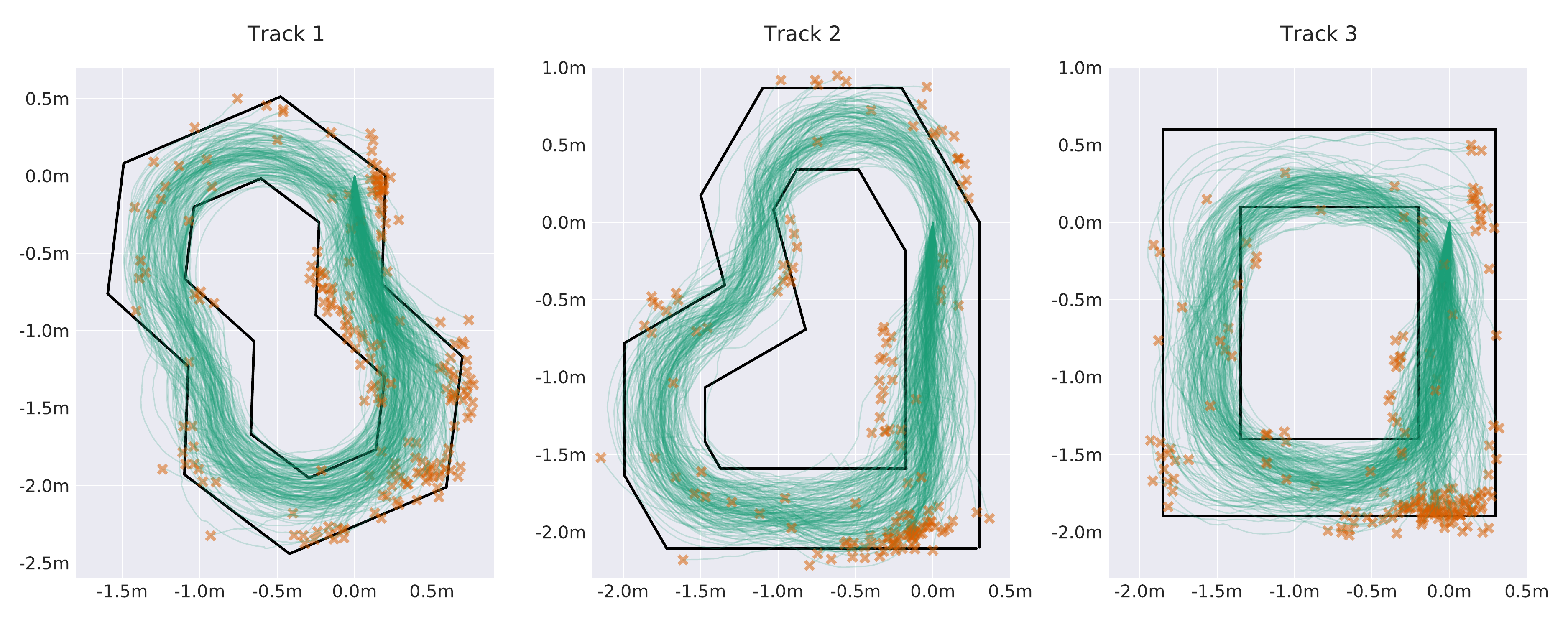}
\caption{The tracks used in our real-world experiments, along with all the trajectories taken by our car during training. The width of the tracks is twice as that of the Donkey car. The car starts from the same position in all episodes. The episode terminates if the car drives off the track. The sparse and noisy disengagement rewards provided to the car are denoted using orange X markers. An episode is successful if the car completes three laps around the track.}
\label{fig:tracks}
\end{figure*}

In this section, we describe our experimental setup for learning to drive a Donkey car from scratch using reinforcement learning. Furthermore, we describe out setup in detail with clear instructions to ensure reproducibility of our real-world experiments.

\subsection{Platform}

\subsubsection{Hardware}


We start with a vanilla Donkey car platform assembled from the official Donkey car starter kit.
The kit is based on the HSP 94186 Brushed RC car, which has been fully tested to support the Donkey car platform. We extend this starter kit with two components:

\begin{itemize}
\item RealSense T265:
%
The throttle control of the car is highly sensitive and the resulting speed of the car is adversely affected by factors such as the friction of the surface, battery voltage, and temperature. In this work, we extend the Donkey car platform with an Intel RealSense T265 and use a PD controller to deal with this issue. The RealSense tracking camera T265 is an off-the-shelf low power device with visual inertial odometry and SLAM capabilities, for use in robotics. We use the T265 for speed estimation and use a PD controller (with parameters $K_p = 0.005$ and $K_d = 0.05$) for throttle so as to maintain the speed set-points provided by an RL agent.



\item Extra battery:
%
We mount an extra 10,000 mAh power bank onto the top cage, to power the Raspberry Pi. The standard Donkey car kit uses the 1100 mAh car battery to also power the Raspberry Pi, which leads to quick battery discharge and very short training times that are not feasible for RL. 

\end{itemize}

\subsubsection{Tracks}

We use three different tracks to ensure that our setup is easily reproducible on all of them and to demonstrate the robustness of our baseline RL agents. We use a different shape for each track. 
The three tracks used in our real-world experiments (including their dimensions) are shown in Fig.~\ref{fig:tracks} and a photo of our setup (with Track 1) is shown in Fig.~\ref{fig:track-photo}.

\subsubsection{Communication}

Our experimental setup consists of the Donkey car and a GPU workstation. The car and the GPU workstation communicate with each other using 
MQTT, a lightweight publish-subscribe messaging queue protocol that is run on top of the TCP/IP stack.

We alternate between local data collection and remote training during learning. We transmit the weights of the policy network from the GPU workstation to the car. The Donkey car uses the donkey Python package to continuously run a loop on the on-board Raspberry Pi 4 computer which collects observations from the camera sensor and RealSense T265, computes the controls to be applied by running the policy network, and applies them to the car. The collected episodic data is transmitted back to the workstation, where it is added to the replay buffer that is used to train the agent.

\subsection{Tasks}


In order to formulate the task of learning to drive around a track as a reinforcement learning problem, we need to define the corresponding Markov Decision Process (MDP).
An MDP consists of:
a set of states $\mathcal{S}$,
a set of actions $\mathcal{A}$,
a transition probability function $s_{t+1} \sim p(s_{t+1} \mid s_t, a_t)$ that represents the probability of transitioning to state $s_{t+1} \in \mathcal{S}$ by taking action $a_t \in \mathcal{A}$ in state $s_t \in \mathcal{S}$ at timestep $t$,
a reward function $R(s_t, a_t)$ that provides a scalar reward for entering state $s_t$ using action $a_t$,
and a discount factor $\gamma \in [0, 1]$ to specify the importance of future rewards.
The goal of reinforcement learning is to learn a policy function $a_t \sim \pi(a_t \mid s_t)$ that maximizes expected cumulative rewards $\mathbb{E} \left[ \sum_{t=0}^\infty \gamma^t R(s_t, a_t) \right]$, such that $a_t \sim \pi(a_t \mid s_t)$ and $s_{t+1} \sim p(s_{t+1} \mid s_t, a_t)$.


We use an episodic formulation of an MDP: each episode starts with the car being at the same point in the track (initial state $s_0$) and is terminated when the car drives off the track.
The transition probability function of the corresponding MDP is defined by our real-world platform. The rest of the MDP components are defined as follows.

\subsubsection{State space} We assume that the state of the system is fully described by observations $o_t$ that the agent receives at every time step $t$.
The observations consist of images from an on-board camera and the speed measurements from RealSense.
We preprocess the images from the camera before feeding them to the agent. We first convert the $120 \times 160$ image to grayscale, crop out the top $40 \times 160$ pixels (consisting of irrelevant surroundings), and finally resize the resulting images to $40 \times 40$ pixels.

\subsubsection{Action space}
The action space in our setup consists of speed and steering. The speed controls from the RL agent are provided as set-points to the PD controller which controls the throttle of the car to attain the set-point speed under varying surfaces and battery conditions.

\subsubsection{Reward function}
\label{sec:rewards}
In this benchmark, we aim to evaluate the data-efficiency, robustness and the control performance of RL algorithms.
We formulate the reward function accordingly.

{\bf Task 1: Learning to steer.} To evaluate data-efficiency, the RL agent is tasked with learning to steer the car moving at a fixed speed and the task is considered successful if the agent can consistently complete three laps around the track. The agent receives a reward of 1 during every time-step until it drives off the track and penalty of -10 when it does so. The maximum total reward that the agent can collect in one episode is 1000, which corresponds to successful completion of the task.
We use a simple mechanism to determine if the car has driven off the track: by thresholding out the darker track from the lighter surrounding and approximating its size. We manually terminate the episode if this mechanism fails.

{\bf Task 2: Generalization.} To evaluate robustness, the RL agents have to learn with little or no hyperparameter tuning in the real-world and demonstrate consistent performance across different tracks. We evaluate robustness by training the agent to solve Task~1 on one of the tracks from Fig.~\ref{fig:tracks} and measuring the agent's performance on two other tracks.

{\bf Task 3: High-speed control.} To evaluate control performance, the RL agent is tasked with controlling both the speed and steering of the car and performance is measured in terms of the average lap time of the agent.
The reward in this case is proportional to the speed of the car and the agent receives a reward of 1 when it is driving at full speed. When the agent drives off the track, it receives penalty of $-10 \times speed$.

\subsection{Steps to Reproduce Our Experiments}

In order to reproduce our experiments, one needs to take the following steps:
\begin{itemize}
    \item Buy a Donkey car starter kit or buy the necessary parts to build one (see \url{https://docs.donkeycar.com/guide/build_hardware/} for all details).
    \item Setup track(s) for training (as shown in Fig.~\ref{fig:tracks}).
    \item Follow instructions in the Donkey car user guide (\url{https://docs.donkeycar.com}) to assemble the car, install the supporting software, and to train an imitation learning baseline. 
    \item Buy a power bank and an Intel RealSense T265, mount them to the car, and connect them to the Raspberry Pi.
    \item Install our supporting software from \url{https://github.com/ari-viitala/donkeycar} and run the training script to exactly reproduce the results of our baseline agents.
\end{itemize}
We also release the complete trajectories of our car during each episode for all training runs, for exact reproducibility (see Fig.~\ref{fig:tracks} for a plot of all trajectories on our three tracks).

\subsection{Simulator}

The Donkey car simulator is based on the Unity game platform and is wrapped in an easily accessible OpenAI Gym interface. We used the Donkey car simulator in our experiments to test the performance of the RL agents to learn Task~1 on randomly generated tracks.

\section{Agents}


\subsection{Imitation learning}
\label{sec:imitation}

Imitation learning involves training a neural network to predict the throttle and steering actions taken by a human. The Donkey car community has tested various network architectures on the platform (see \url{https://docs.donkeycar.com/parts/keras/}) and we use the best performing categorical network in our experiments. The categorical model consists of five convolutional layers followed by two fully-connected layers. The categorical model discretizes the throttle and steering values and the network is trained using categorical cross entropy loss on a manually collected dataset. We collect the data by driving the car around the track for 10K steps, using a Logitecth 710 controller. Detailed instructions to train this model is provided in \url{https://docs.donkeycar.com/guide/train_autopilot/}.

\subsection{Model-free RL}

We use the state-of-the-art soft actor-critic (SAC) \cite{haarnoja2018softa} algorithm as our model-free baseline.
SAC is an off-policy RL algorithm that learns a stochastic actor to maximize an entropy-regularized version of the RL objective that also 
includes
the entropy $\mathcal{H}(\pi(a_t \mid s_t))$ of the learned policy:
\[
\mathbb{E} \left[ \sum_{t=0}^\infty \gamma^t \Big( R(s_t, a_t) + \alpha\mathcal{H}(\pi(a_t \mid s_t)) \Big) \right] \,,
\]
where $\alpha > 0$ is the \emph{temperature} parameter to balance the joint optimization of the cumulative rewards and the entropy.

Learning a policy using raw image observations as states $s_t$ is challenging, as we show in our experiments with the simulator.
Therefore, in the real-world platform we pre-process the high-dimensional image observations by means of a variational autoencoder (VAE) \cite{kingma2013auto}.
The VAE consists of an encoder $q(s_t \mid o_t)$ which encodes observations $o_t$ to create low-dimensional states $s_t$ and a decoder $q(o_t \mid s_t)$ which reconstructs observations $o_t$ from states $s_t$. The encoder and decoder parameters are trained to maximize an objective which can be seen as the sum of a reconstruction loss and a regularization term which tries to keep the state distribution close to the Gaussian distribution with zero mean and an identity covariance matrix. 
We use the states $s_t$ produced by the encoder as the input of the SAC agent.\footnote{The input of SAC is actually a concatenation of the encoder outputs, the speed measurements and five past control signals. We omit this detail in the text for brevity.}

In our experiments, both the encoder and decoder are modeled with convolutional neural networks.
The encoder consists of three $3 \times 3$ convolutional layers and a linear layer to predict the mean and variance parameters of distribution $q(s_t \mid o_t)$. The first convolutional layer has 16 channels with a stride of 2 and others have 32 channels with a stride of 1. The mean parameters produced by the encoder are used as inputs of the SAC networks. The dimensionality of the encoder states $s_t$ is chosen to be 20. The decoder consists of a linear layer and three deconvolutional layers.
The policy and value networks of SAC are fully connected networks with two hidden layers and 64 hidden units in each layer. We use the ReLU non-linearity in all the networks.

We train the VAE and the SAC agent jointly which means that the computational graph built in the optimization procedure contains the encoder and decoder of the VAE and the actor and critic of SAC. Similarly to previous works \cite{tassa2018deepmind, yarats2019improving}, we truncate the gradients from the SAC actor loss to the VAE encoder for stable training.
We use Adam optimizer with a learning rate of 0.0001 for all the networks.
We perform 600 updates of the parameters at the end of each episode using batches of size 128 from the replay buffer.

\subsection{Model-based RL}

Our model-based baseline agent is based on the Dreamer algorithm \cite{hafner2020dream}.
Dreamer summarizes the agent's experience by learning a world model which consists of several components: the representation model $p(s_t \mid s_{t-1},a_{t-1},o_t)$ that encodes observations $o_t$ and actions $a_{t-1}$ to create 
states $s_t$, the observation model $q(o_t\mid s_t)$ that predicts observations $o_t$ from states $s_t$,
the transition model $q(s_t\mid s_{t-1}, a_{t-1})$ that predicts future model states $s_t$ without seeing the corresponding observations $o_t$ and the reward model $q(r_t\mid s_t)$ that predicts the rewards given the model states $s_t$. The world model is trained using past trajectories $\{(a_t,o_t,r_t)\}_{t=k}^{k+L}$ to maximize the variational lower bound, a loss formulated similarly to the VAE loss.

The agent acts in the environment using a policy network $\pi(a_t\mid s_t)$ which is conditioned on the state $s_t$ produced by the representation model $p(s_t \mid s_{t-1},a_{t-1},o_t)$ at every time step $t$.
The policy network is trained using {\it imagined} trajectories $\{(s_\tau,a_\tau,r_\tau)\}_{\tau=t}^{t+H}$ produced by the world model in the following way. First, the representation model computes the initial state $s_t$ from one of the observations from the replay buffer, then the transition and reward models are used to simulate a sequence of future states and rewards. The parameters of the policy network are optimized to maximize the expected return computed using the predicted rewards and the value network, which is the last parameteric component of the Dreamer model.

In our experiments, we use a world model with 30-dimensional states $s_t$. The representation model $p(s_t \mid s_{t-1},a_{t-1},o_t)$ is implemented using a convolutional encoder and a recurrent neural network (RNN) with gated recurrent units (GRU) \cite{cho2014properties}. The RNN processes the previous states $s_{t-1}$ and actions $a_{t-1}$, while the encoder processes the observations $o_t$. Then, the GRU states and the output of the encoder are fed to a multi-layer layer perceptron (MLP) which produces the parameters of the Gaussian distribution $p(s_t \mid s_{t-1},a_{t-1},o_t)$.
The dimensionality of the GRU states is chosen to be 200. The encoder contains four convolutional layers followed by a linear layer to produce a 1024-dimensional vector.
The MLP contains one hidden layer with 300 hidden units. 
The transition model $q(s_t\mid s_{t-1}, a_{t-1})$ is similar to the representation model but it does not use the observations $o_t$ and therefore does not have the encoder.
The reward model, the value model and the policy network are modeled with MLPs with three, three and four hidden layers respectively, each hidden layer having 300 units. The activation function is ELU in all the MLPs and ReLU after the convolutional encoder layers of the representation model.

We update the world model after each episode doing 100 update steps for all the components. The target value network is updated once after each episode. We use the Adam optimizer for training. The learning rate is $6 \cdot 10^{-4}$ for the representation, transition and reward models and $8 \cdot 10^{-5}$ for the value model and the policy network. We clip the gradient norm by 100.

\section{Experiments}
\label{sec:experiments}

In this section, we evaluate our baseline agents on the real-world using the Donkey car platform, and report baseline scores for our proposed L2D benchmark.

\begin{figure}[t]
\centering
\includegraphics[trim=20 15 30 25, clip, width=1.0\linewidth]{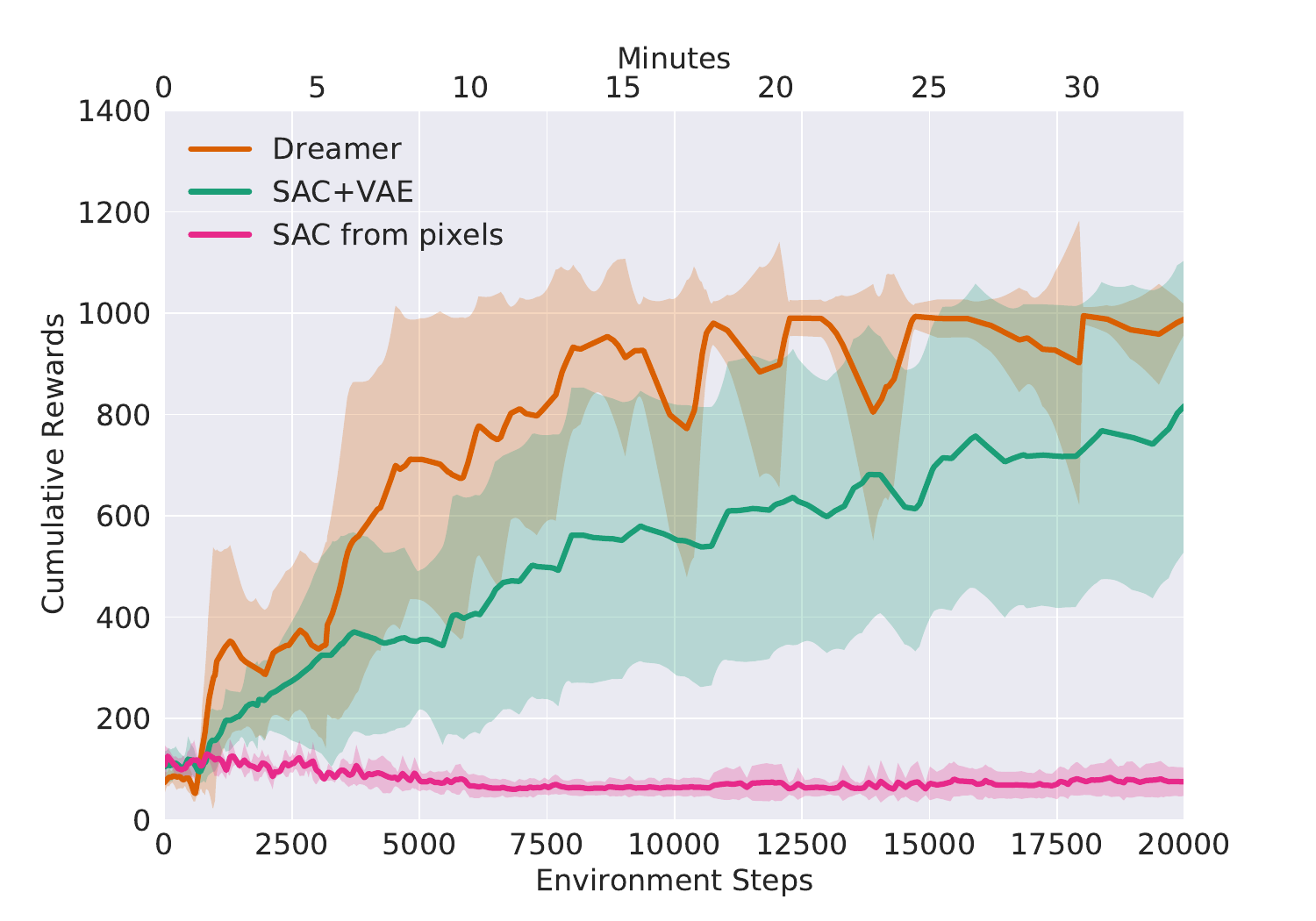}
\caption{
Comparison of our agents on the Donkey simulator. We plot the mean and standard deviation of ten different runs (with randomly generated tracks) for every agent. 
\emph{SAC from pixels} is unable to learn, \emph{SAC+VAE} learns slowly, and \emph{Dreamer} learns robustly in a sample-efficient manner.
}
\label{fig:results-simulator}
\end{figure}

\begin{figure*}[t]
\centering
\includegraphics[trim=20 20 20 20, clip, width=1.0\linewidth]{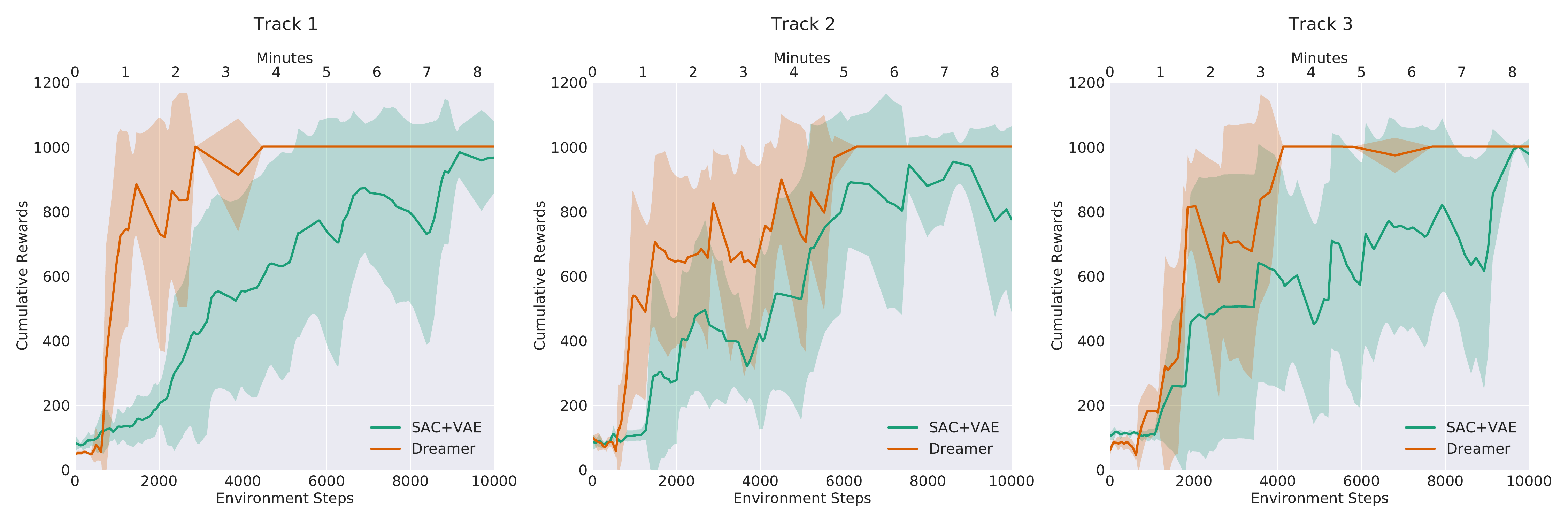}
\caption{
Evaluation of baseline agents on the proposed benchmark with three real-world tracks. We plot the mean and standard deviation of five different runs for every agent. \emph{Dreamer} learns from sparse and noisy rewards in a very sample-efficient and robust manner to achieve the maximum return of 1000 within merely 5 minutes of driving, in all tracks. \emph{SAC+VAE} is also able to learn, but requires almost twice as many disengagements as \emph{Dreamer}.
}
\label{fig:results-real}
\end{figure*}

\subsection{Task 1: Learning to Steer}

We first aim to evaluate the sample-efficiency and robustness of three learning agents: 1) \emph{SAC from pixels}, 2) \emph{SAC+VAE}, and 3) \emph{Dreamer}, on Task~1, using randomly generated roads in the Donkey car simulator. 
\emph{SAC from pixels} uses images as inputs of SAC (and thus does not have a VAE) while \emph{SAC+VAE} uses the output of the VAE encoder as inputs of SAC.
We use a fixed throttle of 0.6 and a control frequency of 10Hz.
The results of our experiments are shown in Fig.~\ref{fig:results-simulator}. The maximum cumulative reward that an agent can achieve in this task is 1000. While \emph{SAC from pixels} is unable to learn anything, \emph{SAC+VAE} is able to achieve the maximal performance most of the time but fails to do so in some tracks. \emph{Dreamer} algorithm performs the best, learning robustly and significantly faster than SAC, on all simulator experiments. \emph{SAC+VAE} requires almost twice as many disengagements to learn than \emph{Dreamer}.

\begin{table}[t]
    \centering
    \caption{Results of our experiments on Task 2: generalization}
    \label{tab:genralization}
    \begin{tabular}{c|c|cc}
    \toprule
        Train & Test & SAC+VAE & Dreamer \\
    \midrule
        \multirow{3}{*}{Track 1} 
        & Track 1 & $100.0\pm00.0\%$ & $100.0\pm00.0\%$ \\     
        & Track 2 & $100.0\pm00.0\%$ & $100.0\pm00.0\%$ \\
        & Track 3 & $100.0\pm00.0\%$ & $100.0\pm00.0\%$ \\
    \midrule
        \multirow{3}{*}{Track 2} 
        & Track 1 & $100.0\pm00.0\%$ & $100.0\pm00.0\%$ \\
        & Track 2 & $50.58\pm30.6\%$ & $89.65\pm20.7\%$ \\
        & Track 3 & $94.78\pm07.4\%$ & $47.79\pm30.3\%$ \\
    \midrule
        \multirow{3}{*}{Track 3} 
        & Track 1 & $21.70\pm21.6\%$ & $98.92\pm01.5\%$ \\     
        & Track 2 & $22.32\pm00.2\%$ & $24.98\pm00.2\%$ \\
        & Track 3 & $100.0\pm00.0\%$ & $100.0\pm00.0\%$ \\
    \bottomrule
    \end{tabular}
\end{table}

\begin{table}[t]
    \centering
    \caption{Results of our experiments on Task 3: high-speed control}
    \label{tab:speed}
    \begin{tabular}{l|cccc}
    \toprule
        Agent & Human & Imitation & SAC+VAE & Dreamer \\
    \midrule
        Lap time & 5.82~s & 6.62~s & \textbf{4.92~s} & 5.79~s \\
    \bottomrule
    \end{tabular}
\end{table}

We consider the \emph{SAC+VAE} and \emph{Dreamer} agents that are able to learn to drive in the simulator and evaluate their performance on the physical platform (see Task~1 described in Section~\ref{sec:rewards}).
We use the PD controller to maintain a fixed speed of 0.5 m/s. The agents control the car at a frequency of 20Hz (that is, each environment corresponds to 0.05 seconds). The results of our experiments on three different tracks are shown in Fig.~\ref{fig:results-real}. Similar to our results on the simulator, the \emph{Dreamer} agent learns in a very sample-efficient and robust manner on all the three tracks, outperforming the \emph{SAC+VAE} agent that requires almost twice as many disengagements to learn. The \emph{Dreamer} agent can learn to drive around all tracks within just 5 minutes of experience.

\subsection{Task 2: Generalization}

We take agents trained on one track and test their performance on the other tracks. The mean and std of the agents over five episodes (as a percentage of the maximum achievable return of 1000) is shown in Table~\ref{tab:genralization}. 
Environmental conditions such as ambient lighting during training and testing times were different and we observe more variance in the performance of our agents.
\emph{Dreamer} generalizes better than \emph{SAC+VAE}. While agents trained on Track 1 are able to generalize to all other tracks, agents trained on Track 3 are unable to generalize to other tracks (since Track 3 only consists of sharp right-angle bends).
Agents trained on Track 2 perform better on Track 1, which is easier to drive on.

\subsection{Task 3: High-Speed Control}

To test the driving performance of RL agents, we train them on Track~1 to control both the speed and steering of the car (see Task~3 described in Section~\ref{sec:rewards}).
We initialize the weights of agents trained on Task 1, and train them till convergence on Task 3. 
We compare the average lap times over five laps (from a flying start) of best performing RL agents with a well-tested imitation learning agent (described in Section~\ref{sec:imitation}) and a human operator in Table~\ref{tab:speed}. We manually drive the car as fast as possible using a Logitech 710 joystick to achieve an average lap time of 5.82~s. Imitation learning is able to closely match this performance (6.62~s). We demonstrate that reinforcement learning agents are able to outperform these baselines, learning from just sparse and noisy disengagement signals. Although \emph{SAC+VAE} requires more samples to learn, it is able to drive the fastest (4.92~s), outperforming \emph{Dreamer}, which is also able to drive faster (5.79~s) than a human.

\section{Conclusion}
\label{sec:conclusion}

In this paper, we propose Learning to Drive (L2D), a low-cost benchmark for research in real-world RL. We comprehensively describe the experimental setup so that the results can be reproduced exactly. We demonstrate that existing RL algorithms can learn to drive the car from scratch in less than five minutes of interaction. We also show that RL agents trained with sparse and noisy disengagement signals can learn to drive the car faster than an imitation-learning agent and a human operator.
See \url{https://sites.google.com/view/donkeycar} for a supplementary video containing a complete training session of Dreamer on Track~1, evaluation of the trained policy on other tracks, and comparison of all agents on high-speed control.

The proposed benchmark has great potential to be extended to more challenging RL tasks in the future. We consider using larger outdoor tracks, navigating to target locations and scenarios involving multiple cars.








\bibliographystyle{IEEEtran}
\bibliography{IEEEabrv,root}

\begin{thebibliography}{10}
\providecommand{\url}[1]{#1}
\csname url@rmstyle\endcsname
\providecommand{\newblock}{\relax}
\providecommand{\bibinfo}[2]{#2}
\providecommand\BIBentrySTDinterwordspacing{\spaceskip=0pt\relax}
\providecommand\BIBentryALTinterwordstretchfactor{4}
\providecommand\BIBentryALTinterwordspacing{\spaceskip=\fontdimen2\font plus
\BIBentryALTinterwordstretchfactor\fontdimen3\font minus
  \fontdimen4\font\relax}
\providecommand\BIBforeignlanguage[2]{{%
\expandafter\ifx\csname l@#1\endcsname\relax
\typeout{** WARNING: IEEEtran.bst: No hyphenation pattern has been}%
\typeout{** loaded for the language `#1'. Using the pattern for}%
\typeout{** the default language instead.}%
\else
\language=\csname l@#1\endcsname
\fi
#2}}

\bibitem{duan2016benchmarking}
Y.~Duan, X.~Chen, R.~Houthooft, J.~Schulman, and P.~Abbeel, ``Benchmarking deep
  reinforcement learning for continuous control,'' in \emph{International
  Conference on Machine Learning}, 2016, pp. 1329--1338.

\bibitem{islam2017reproducibility}
R.~Islam, P.~Henderson, M.~Gomrokchi, and D.~Precup, ``Reproducibility of
  benchmarked deep reinforcement learning tasks for continuous control,''
  \emph{arXiv preprint arXiv:1708.04133}, 2017.

\bibitem{henderson2018deep}
P.~Henderson, R.~Islam, P.~Bachman, J.~Pineau, D.~Precup, and D.~Meger, ``Deep
  reinforcement learning that matters,'' in \emph{AAAI}, 2018.

\bibitem{mania2018simple}
H.~Mania, A.~Guy, and B.~Recht, ``Simple random search provides a competitive
  approach to reinforcement learning,'' \emph{arXiv preprint arXiv:1803.07055},
  2018.

\bibitem{haarnoja2018softb}
T.~Haarnoja, A.~Zhou, K.~Hartikainen, G.~Tucker, S.~Ha, J.~Tan, V.~Kumar,
  H.~Zhu, A.~Gupta, P.~Abbeel, \emph{et~al.}, ``Soft actor-critic algorithms
  and applications,'' \emph{arXiv preprint arXiv:1812.05905}, 2018.

\bibitem{hafner2020dream}
D.~Hafner, T.~Lillicrap, J.~Ba, and M.~Norouzi, ``Dream to control: Learning
  behaviors by latent imagination,'' in \emph{International Conference on
  Learning Representations}, 2020.

\bibitem{brockman2016openai}
G.~Brockman, V.~Cheung, L.~Pettersson, J.~Schneider, J.~Schulman, J.~Tang, and
  W.~Zaremba, ``{OpenAI Gym},'' \emph{arXiv preprint arXiv:1606.01540}, 2016.

\bibitem{tassa2018deepmind}
Y.~Tassa, Y.~Doron, A.~Muldal, T.~Erez, Y.~Li, D.~d.~L. Casas, D.~Budden,
  A.~Abdolmaleki, J.~Merel, A.~Lefrancq, \emph{et~al.}, ``Deepmind control
  suite,'' \emph{arXiv preprint arXiv:1801.00690}, 2018.

\bibitem{akkaya2019solving}
I.~Akkaya, M.~Andrychowicz, M.~Chociej, M.~Litwin, B.~McGrew, A.~Petron,
  A.~Paino, M.~Plappert, G.~Powell, R.~Ribas, \emph{et~al.}, ``Solving rubik's
  cube with a robot hand,'' \emph{arXiv preprint arXiv:1910.07113}, 2019.

\bibitem{bewley2019learning}
A.~Bewley, J.~Rigley, Y.~Liu, J.~Hawke, R.~Shen, V.-D. Lam, and A.~Kendall,
  ``Learning to drive from simulation without real world labels,'' in
  \emph{2019 International Conference on Robotics and Automation (ICRA)}.\hskip
  1em plus 0.5em minus 0.4em\relax IEEE, 2019, pp. 4818--4824.

\bibitem{goodfellow2016}
I.~Goodfellow, Y.~Bengio, and A.~Courville, \emph{Deep Learning}.\hskip 1em
  plus 0.5em minus 0.4em\relax MIT Press, 2016,
  \url{http://www.deeplearningbook.org}.

\bibitem{imagenet}
J.~Deng, W.~Dong, R.~Socher, L.-J. Li, K.~Li, and L.~Fei-Fei, ``{ImageNet: A
  Large-Scale Hierarchical Image Database},'' in \emph{CVPR}, 2009.

\bibitem{kendall2019learning}
A.~Kendall, J.~Hawke, D.~Janz, P.~Mazur, D.~Reda, J.-M. Allen, V.-D. Lam,
  A.~Bewley, and A.~Shah, ``Learning to drive in a day,'' in \emph{2019
  International Conference on Robotics and Automation (ICRA)}.\hskip 1em plus
  0.5em minus 0.4em\relax IEEE, 2019, pp. 8248--8254.

\bibitem{donkeycar}
``Donkey car,'' \url{https://www.donkeycar.com/}, accessed: 2020-07-27.

\bibitem{dulac2019challenges}
G.~Dulac-Arnold, D.~Mankowitz, and T.~Hester, ``Challenges of real-world
  reinforcement learning,'' \emph{arXiv preprint arXiv:1904.12901}, 2019.

\bibitem{yarats2019improving}
D.~Yarats, A.~Zhang, I.~Kostrikov, B.~Amos, J.~Pineau, and R.~Fergus,
  ``Improving sample efficiency in model-free reinforcement learning from
  images,'' \emph{arXiv preprint arXiv:1910.01741}, 2019.

\bibitem{ahn2020robel}
M.~Ahn, H.~Zhu, K.~Hartikainen, H.~Ponte, A.~Gupta, S.~Levine, and V.~Kumar,
  ``Robel: Robotics benchmarks for learning with low-cost robots,'' in
  \emph{Conference on Robot Learning}.\hskip 1em plus 0.5em minus 0.4em\relax
  PMLR, 2020, pp. 1300--1313.

\bibitem{yang2019replab}
B.~Yang, D.~Jayaraman, J.~Zhang, and S.~Levine, ``Replab: A reproducible
  low-cost arm benchmark for robotic learning,'' in \emph{2019 International
  Conference on Robotics and Automation (ICRA)}.\hskip 1em plus 0.5em minus
  0.4em\relax IEEE, 2019, pp. 8691--8697.

\bibitem{balaji2019deepracer}
B.~Balaji, S.~Mallya, S.~Genc, S.~Gupta, L.~Dirac, V.~Khare, G.~Roy, T.~Sun,
  Y.~Tao, B.~Townsend, \emph{et~al.}, ``Deepracer: Educational autonomous
  racing platform for experimentation with sim2real reinforcement learning,''
  \emph{arXiv preprint arXiv:1911.01562}, 2019.

\bibitem{paull2017duckietown}
L.~Paull, J.~Tani, H.~Ahn, J.~Alonso-Mora, L.~Carlone, M.~Cap, Y.~F. Chen,
  C.~Choi, J.~Dusek, Y.~Fang, \emph{et~al.}, ``Duckietown: an open, inexpensive
  and flexible platform for autonomy education and research,'' in \emph{2017
  IEEE International Conference on Robotics and Automation (ICRA)}.\hskip 1em
  plus 0.5em minus 0.4em\relax IEEE, 2017, pp. 1497--1504.

\bibitem{gorjup2020mushr}
G.~{Gorjup} and M.~{Liarokapis}, ``A low-cost, open-source, robotic airship for
  education and research,'' \emph{IEEE Access}, vol.~8, pp. 70\,713--70\,721,
  2020.

\bibitem{robomaster}
``{DJI Robomaster S1},'' \url{https://www.dji.com/fi/robomaster-s1}, accessed:
  2020-10-27.

\bibitem{jetbot}
``{NVIDIA JetBot},'' \url{https://github.com/nvidia-ai-iot/jetbot}, accessed:
  2020-10-27.

\bibitem{dosovitskiy2017carla}
A.~Dosovitskiy, G.~Ros, F.~Codevilla, A.~Lopez, and V.~Koltun, ``Carla: An open
  urban driving simulator,'' in \emph{Conference on Robot Learning}, 2017, pp.
  1--16.

\bibitem{singh2019}
A.~Singh, L.~Yang, K.~Hartikainen, C.~Finn, and S.~Levine, ``End-to-end robotic
  reinforcement learning without reward engineering,'' \emph{Robotics: Science
  and Systems}, 2019.

\bibitem{ha2020learning}
S.~Ha, P.~Xu, Z.~Tan, S.~Levine, and J.~Tan, ``Learning to walk in the real
  world with minimal human effort,'' \emph{arXiv preprint arXiv:2002.08550},
  2020.

\bibitem{yang2020data}
Y.~Yang, K.~Caluwaerts, A.~Iscen, T.~Zhang, J.~Tan, and V.~Sindhwani, ``Data
  efficient reinforcement learning for legged robots,'' in \emph{Conference on
  Robot Learning}.\hskip 1em plus 0.5em minus 0.4em\relax PMLR, 2020, pp.
  1--10.

\bibitem{haarnoja2018learning}
T.~Haarnoja, S.~Ha, A.~Zhou, J.~Tan, G.~Tucker, and S.~Levine, ``Learning to
  walk via deep reinforcement learning,'' \emph{arXiv preprint
  arXiv:1812.11103}, 2018.

\bibitem{zhu2019ingredients}
H.~Zhu, J.~Yu, A.~Gupta, D.~Shah, K.~Hartikainen, A.~Singh, V.~Kumar, and
  S.~Levine, ``The ingredients of real world robotic reinforcement learning,''
  in \emph{International Conference on Learning Representations}, 2019.

\bibitem{lillicrap2016continuous}
T.~P. Lillicrap, J.~J. Hunt, A.~Pritzel, N.~Heess, T.~Erez, Y.~Tassa,
  D.~Silver, and D.~Wierstra, ``Continuous control with deep reinforcement
  learning.'' in \emph{ICLR}, 2016.

\bibitem{haarnoja2018softa}
T.~Haarnoja, A.~Zhou, P.~Abbeel, and S.~Levine, ``Soft actor-critic: Off-policy
  maximum entropy deep reinforcement learning with a stochastic actor,'' in
  \emph{International Conference on Machine Learning}, 2018, pp. 1861--1870.

\bibitem{shelhamer2016loss}
E.~Shelhamer, P.~Mahmoudieh, M.~Argus, and T.~Darrell, ``Loss is its own
  reward: Self-supervision for reinforcement learning,'' \emph{arXiv preprint
  arXiv:1612.07307}, 2016.

\bibitem{higgins2017darla}
I.~Higgins, A.~Pal, A.~Rusu, L.~Matthey, C.~Burgess, A.~Pritzel, M.~Botvinick,
  C.~Blundell, and A.~Lerchner, ``Darla: Improving zero-shot transfer in
  reinforcement learning,'' in \emph{International Conference on Machine
  Learning}, 2017, pp. 1480--1490.

\bibitem{nair2018visual}
A.~V. Nair, V.~Pong, M.~Dalal, S.~Bahl, S.~Lin, and S.~Levine, ``Visual
  reinforcement learning with imagined goals,'' in \emph{Advances in Neural
  Information Processing Systems}, 2018, pp. 9191--9200.

\bibitem{lange2010deep}
S.~Lange and M.~A. Riedmiller, ``Deep learning of visual control policies,'' in
  \emph{ESANN}, 2010.

\bibitem{lambert2019low}
N.~O. Lambert, D.~S. Drew, J.~Yaconelli, S.~Levine, R.~Calandra, and K.~S.
  Pister, ``Low-level control of a quadrotor with deep model-based
  reinforcement learning,'' \emph{IEEE Robotics and Automation Letters},
  vol.~4, no.~4, pp. 4224--4230, 2019.

\bibitem{becker2020learning}
P.~Becker-Ehmck, M.~Karl, J.~Peters, and P.~van~der Smagt, ``Learning to fly
  via deep model-based reinforcement learning,'' \emph{arXiv preprint
  arXiv:2003.08876}, 2020.

\bibitem{ebert2018visual}
F.~Ebert, C.~Finn, S.~Dasari, A.~Xie, A.~Lee, and S.~Levine, ``Visual
  foresight: Model-based deep reinforcement learning for vision-based robotic
  control,'' \emph{arXiv preprint arXiv:1812.00568}, 2018.

\bibitem{zhang2019solar}
M.~Zhang, S.~Vikram, L.~Smith, P.~Abbeel, M.~Johnson, and S.~Levine, ``Solar:
  Deep structured representations for model-based reinforcement learning,'' in
  \emph{International Conference on Machine Learning}.\hskip 1em plus 0.5em
  minus 0.4em\relax PMLR, 2019, pp. 7444--7453.

\bibitem{kingma2013auto}
D.~P. Kingma and M.~Welling, ``Auto-encoding variational bayes,'' \emph{arXiv
  preprint arXiv:1312.6114}, 2013.

\bibitem{cho2014properties}
K.~Cho, B.~van Merrienboer, D.~Bahdanau, and Y.~Bengio, ``On the properties of
  neural machine translation: Encoder-decoder approaches,'' in \emph{Eighth
  Workshop on Syntax, Semantics and Structure in Statistical Translation
  (SSST-8)}, 2014.

\end{thebibliography}

\end{document}